\PassOptionsToPackage{table,dvipsnames}{xcolor}
\documentclass[10pt,twocolumn,letterpaper]{article}

\usepackage{iccv}              

\definecolor{iccvblue}{rgb}{0.21,0.49,0.74}
\usepackage[pagebackref,breaklinks,colorlinks,allcolors=iccvblue]{hyperref}

\usepackage{graphicx}
\usepackage{float}
\usepackage{adjustbox}
\usepackage{changepage}
\usepackage{balance}
\usepackage{lipsum}

\usepackage{array}
\usepackage{placeins}
\usepackage{booktabs}
\usepackage{multirow}
\usepackage{makecell}
\usepackage{tabularx}
\usepackage{tabularray}
\usepackage{colortbl}
\usepackage{rotating}
\usepackage{caption}
\usepackage[font=small,skip=2pt]{caption}
\usepackage{xcolor}
\usepackage{pifont}
\usepackage{ragged2e} 

\usepackage[most]{tcolorbox}

\definecolor{headerblue}{RGB}{34,85,160}
\definecolor{rowgray}{RGB}{245,245,245}
\definecolor{goodgreen}{RGB}{200,255,200}
\definecolor{warnred}{RGB}{255,220,220}

\newcommand{\xmark}{\ding{55}} 
\tcbset{
  promptbox/.style={
    colback=gray!5,
    colframe=black!30,
    width=\columnwidth,
    boxrule=0.3pt,
    arc=1mm,
    left=4pt,
    right=4pt,
    top=4pt,
    bottom=4pt,
    boxsep=1pt
  }
}

\definecolor{lightgray}{gray}{0.95}
\definecolor{lightblue}{rgb}{0.90,0.95,1.0}
\definecolor{domaincolor}{RGB}{255,250,230}     
\definecolor{foundationcolor}{RGB}{235,245,255} 
\definecolor{fusioncolor}{RGB}{240,255,240}     

\renewcommand{\arraystretch}{1.0}
\def\paperID{*****} 
\def\confName{ICCV}
\def\confYear{2025}

\title{Foundation versus Domain-specific Models:\\ Performance Comparison, Fusion, and Explainability in Face Recognition}

\vspace{-1em}
\author{
Redwan Sony$^{*}$, Parisa Farmanifard, Arun Ross$^{*}$, Anil K. Jain\\
Michigan State University\\
\vspace{-0.5em}
{\tt\small \{sonymd, farmanif, rossarun, jain\}@msu.edu}
}
\DeclareUnicodeCharacter{202F}{\,}

\begin{document}
\twocolumn[{%
    \maketitle
    \vspace{-3em}
    \begin{center}
        \renewcommand{\thefigure}{1}  
        \refstepcounter{figure} 
        \includegraphics[width=0.95\textwidth]{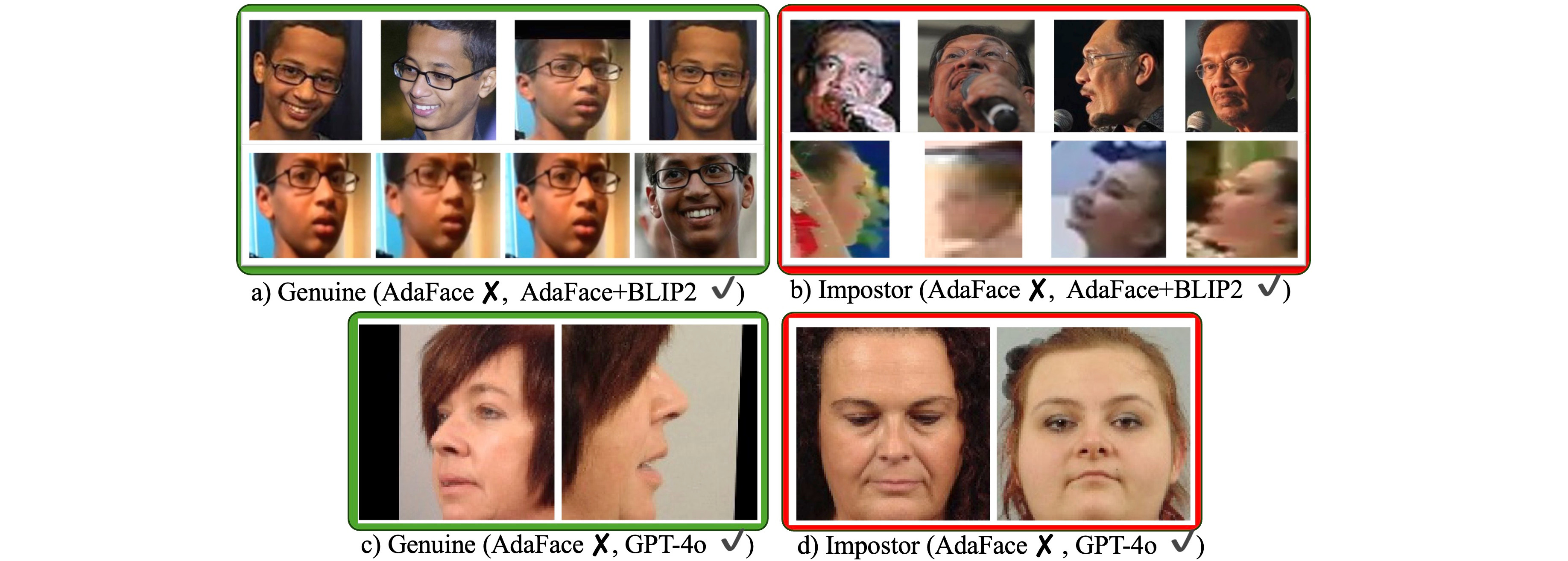}\\[4pt]
        \small
        \vspace{-0.6em}
        \textbf{Figure~\thefigure.} 
        Example face pairs from IJB-B~\cite{whitelam2017ijbb} and IJB-S~\cite{kalka2018ijbs} datasets misclassified by AdaFace~\cite{kim2022adaface} but correctly classified after the fusion of AdaFace and BLIP2~\cite{li2023blip2} or AdaFace and GPT-4o~\cite{openai2024gpt4o}.
        \label{fig:visual_abstract}
    \end{center}
}]

\renewcommand{\thefootnote}{\fnsymbol{footnote}}
\footnotetext[1]{Corresponding authors}
\renewcommand{\thefootnote}{\arabic{footnote}}

\begin{abstract}
In this paper, we address the following question: How do generic foundation models (e.g., CLIP, BLIP, GPT-4o, Grok-4) compare against a domain-specific face recognition model (viz., AdaFace or ArcFace) on the face recognition task? Through a series of experiments involving several foundation models and benchmark datasets, we report the following findings: (a) In all face benchmark datasets considered, domain-specific models outperformed zero-shot foundation models. (b) The performance of zero-shot generic foundation models improved on over-segmented face images compared to tightly cropped faces, thereby suggesting the importance of contextual clues. (c) A simple score-level fusion of a foundation model with a domain-specific face recognition model improved the accuracy at low false match rates. (d) Foundation models, such as GPT-4o and Grok-4, are able to provide explainability to the face recognition pipeline. In some instances, foundation models are even able to resolve low-confidence decisions made by AdaFace,  thereby reiterating the importance of combining domain-specific face recognition models with generic foundation models in a judicious manner.  
\end{abstract}

\vspace{-1em}
\section{Introduction}
\label{sec:intro}
\vspace{-0.5em}
Automated Face Recognition (FR) is an important biometric technology that uses face images or videos to recognize an individual~\cite{jain2025biometrics}. An effective FR model should exhibit extremely high accuracy, robustness to variations (e.g., pose, illumination, and expression), scalability across large populations, compact template size, low latency, security against attacks, and interpretability. In the most recent NIST face recognition  technology evaluations~\cite{nist2025frte11}, the leading FR model had a False Non-Match Rate (FNMR) of $0.0007$ at a False Match Rate (FMR) of $0.0001$ in the constrained cooperative scenario (``VISA"). In the unconstrained non-cooperative scenario (``BORDER-KIOSK"), the top-performing FR model achieved an FNMR of $0.0338$ at an FMR of $0.00001$. This remarkable level of accuracy compared to the accuracies in early NIST evaluations has been possible due to the use of deep neural network models trained on huge amounts of face training data. We refer to these models as \textit{domain-specific FR  models} since they have been exclusively trained for the FR domain. Notable among them and available as open source are AdaFace~\cite{kim2022adaface}, ArcFace~\cite{deng2019arcface}, CosFace~\cite{Wang2018Cosface:Recognition}, MagFace~\cite{meng2021magface}, FaceNet~\cite{schroff2015facenet}, and SphereFace~\cite{liu2017sphereface}. A comprehensive historical and technical review of FR frameworks, models, datasets, and benchmarks can be found in Kim~\textit{et al.}~\cite{kim202550yearsautomatedface}.

\begin{table}[t]
    \centering
    \caption{\small Comparison between Face Recognition (FR) and  Foundation Models}
    \small
    \renewcommand{\arraystretch}{1.2}
    \rowcolors{2}{lightgray}{white}
    \resizebox{\linewidth}{!}{%
    \begin{tabular}{p{4.1cm}|p{4.1cm}}
        \Xhline{1px}
        \rowcolor{lightblue}
        \makecell[c]{\textbf{Face Recognition (FR) Models}}
        & \makecell[c]{\textbf{Foundation Models }} \\
        \Xhline{1px}
        Trained on only face images & Trained on generic images/text \\
        Exclusively used for FR tasks & Can be used for multiple tasks \\
        Supervised training & Unsupervised training \\
        High accuracy on FR tasks & Lower accuracy on FR tasks \\
        Low inference time & Higher inference time \\
        Limited explainability & Provide textual explanations \\\Xhline{1px}
    \end{tabular}}
    \label{tab:model_coparison}
\end{table}

A \textit{foundation model} is a large-scale model trained on huge quantities of diverse and broad domain data, often spanning multiple modalities (e.g., image, video, text, audio), with the goal of enabling adaptability across a wide range of downstream tasks with minimal fine-tuning~\cite{bommasani2022opportunitiesrisksfoundationmodels}. In particular, vision-language foundation models are becoming very prevalent. These models are typically pre-trained using self-supervised or unsupervised learning objectives that involve aligning visual features with corresponding textual embeddings. Such text-guided learning strategies empower the vision encoders within foundation models to capture semantically rich and transferable visual representations, making them highly effective feature extractors across multiple domains for a variety of tasks~\cite{clip_radford2021learning, Cherti_2023_CVPR_OPENCLIP, li2022blip, jia2021scaling_align}. While we collectively refer to them as foundation models, they fall into two broader categories: a) \textit{open-source }foundation models with accessible vision backbones enabling access to the embedding of the input image, and b) \textit{closed-source} foundation models that can only be accessed through their application programming interface (API).\footnote{The use of APIs, which include opt-out provisions, ensures that uploaded face data is not retained or reused under the service agreement, thereby mitigating privacy risks.} Examples of open-source foundation models include CLIP~\cite{clip_radford2021learning}, OpenCLIP~\cite{Cherti_2023_CVPR_OPENCLIP}, BLIP~\cite{li2022blip, li2023blip2}, DINO~\cite{caron2021dino}, SAM~\cite{mazurowski2023segment, ravi2025sam2}, LLaVA~\cite{llava15, liu2024llavanext}, DeepSeek~\cite{deepseek2024vl, wu2024deepseek_vl2, dai2024deepseekmoe}, InternVL~\cite{chen2024internvl, zhu2025internvl3}, ALIGN~\cite{jia2021scaling_align}, and Kosmos~\cite{peng2023kosmos}. Examples of closed-source foundation models include GPT-4o~\cite{openai2024gpt4o}, Gemini~\cite{geminiteam2025geminifamilyhighlycapable} and Grok-4~\cite{xai2025grok}. Table~\ref{tab:model_coparison} shows a basic comparison between FR models and foundation models.  

In this paper, we raise the following questions.
\begin{enumerate}
\itemsep0em 
\item What are the benefits of using generic foundation models for face recognition? 
\item Can foundation models be used alongside FR models to further improve face recognition accuracy?
\item How well do foundation models perform in providing explainability to the face recognition task? 
\end{enumerate}

In order to investigate these questions, we conducted an extensive study involving thirty-five open-source foundation models (corresponding to fourteen families), three closed-source foundation models, and five FR models (corresponding to two families) on five benchmark datasets.\footnote{For brevity, we report performance on just a subset of these models. The complete set of results can be found in the supplementary materials~\href{https://github.com/iPRoBe-lab/AFRvsVLM/blob/supplementary/README.md}{here.}}

\begin{figure}[t!]
    \centering
    \includegraphics[width=0.99\linewidth]{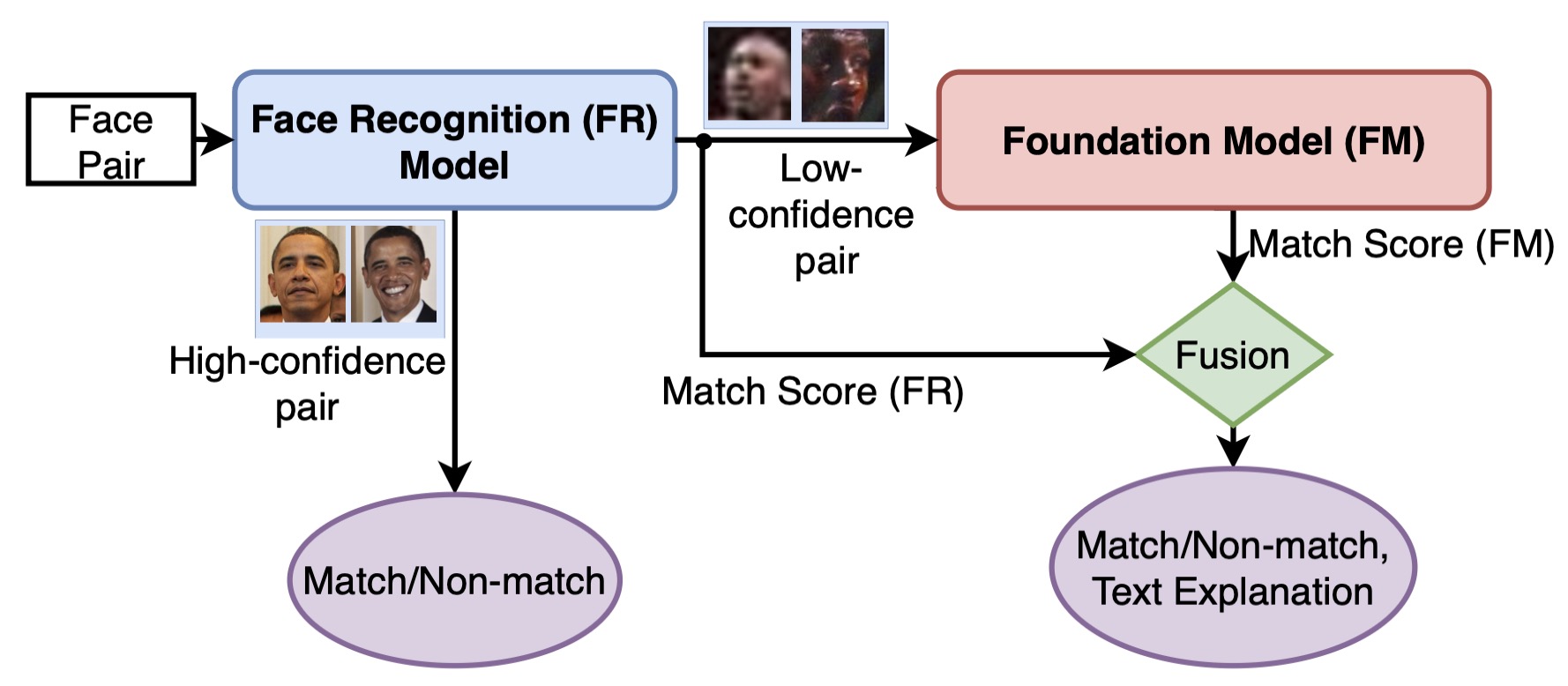}
        \caption{Fusion of face recognition model with foundation model to generate similarity scores and textual explanations.} 
    \label{fig:fusion_diagram}
\end{figure}

Our findings are as follows:
\begin{enumerate}
\itemsep0em
    \item FR models are vastly superior over zero-shot foundation models on the task of face recognition.
    \item The accuracy of a foundation model improves when the face images are {\em loosely} cropped from the background. This suggests the importance of contextual information in the case of foundation models. 
    \item Score-level fusion of foundation model with FR model improves the face recognition accuracy (Figure~\ref{fig:visual_abstract}). 
    \item Foundation models, such as GPT-4o, can be used to provide reasons for a match or a non-match.
    \item Foundation models can also be used to resolve low-confidence decisions rendered by FR models (Figure~\ref{fig:fusion_diagram}).   
\end{enumerate}

In summary, this work demonstrates the benefits of using foundation models in conjunction with FR models to advance face recognition.

\section{Related Work}
\vspace{-0.5em}
Foundation models have been explored for different biometric tasks such as face recognition~\cite{sony2025benchmarking, chettaoui2024froundation}, attribute prediction~\cite{sony2025benchmarking}, iris segmentation~\cite{farmanifard2024iris}, iris recognition~\cite{farmanifard2024chatgpt}, presentation attack detection~\cite{tapia2025towards}, and morph detection~\cite{ozgur2025foundpadfoundationmodelsreloaded, komaty2025exploringchatgptfacepresentation}.
Recent work has explored the use of foundation models in face recognition and related tasks. Bhat et al.~\cite{bhat2023clipface} evaluated CLIP~\cite{clip_radford2021learning} and OpenCLIP~\cite{Cherti_2023_CVPR_OPENCLIP} as zero-shot face recognizers, highlighting their potential despite lacking domain-specific training. DeAndres-Tame et al.~\cite{deandres2024good} employed GPT-4~\cite{openai2023gpt4} for face verification and attribute extraction, emphasizing its ability to provide decision rationales. Hassanpour et al.~\cite{hassanpour2024chatgpt} investigated the performance of ChatGPT~\cite{openai2023gpt4}  on face recognition, age estimation, and gender classification, revealing privacy risks associated with prompt-based querying. Chettaoui et al.~\cite{chettaoui2024froundation} demonstrated that fine-tuning a pre-trained foundation model  yields superior face recognition performance compared to training from scratch in  limited data scenarios. Otroshi et al.~\cite{fmbiometrics2025survey} presented a comprehensive survey on foundation models, categorizing them into four primary types: large language models, vision-language models, audio-language models, and fully multimodal models. Narayan et al.~\cite{narayan2025facexbench} introduced FaceXBench, a large-scale visual question answering style benchmark for facial understanding, and identified reasoning limitations in LLMs. Sony et al.~\cite{sony2025benchmarking} benchmarked 41 foundation models across various biometric tasks including face recognition, morph attack and DeepFake detection, reporting strong zero or few-shot performance across tasks. While prior work has examined either  zero-shot performance or explainability with foundation models, none of these studies provide direct, large-scale comparison with SoTA FR systems. We present a systematic comparison of multiple foundation models against SoTA FR baselines, viz., AdaFace~\cite{kim2022adaface} and ArcFace~\cite{deng2019arcface}, across several benchmark datasets. We also visualize  failure cases, fuse FR models with foundation models to improve performance on challenging cases, and  demonstrate how foundation models can provide textual explanations, bridging the gap between accuracy and explainability.

\begin{figure}[b]
    \centering
    \includegraphics[width=\linewidth]{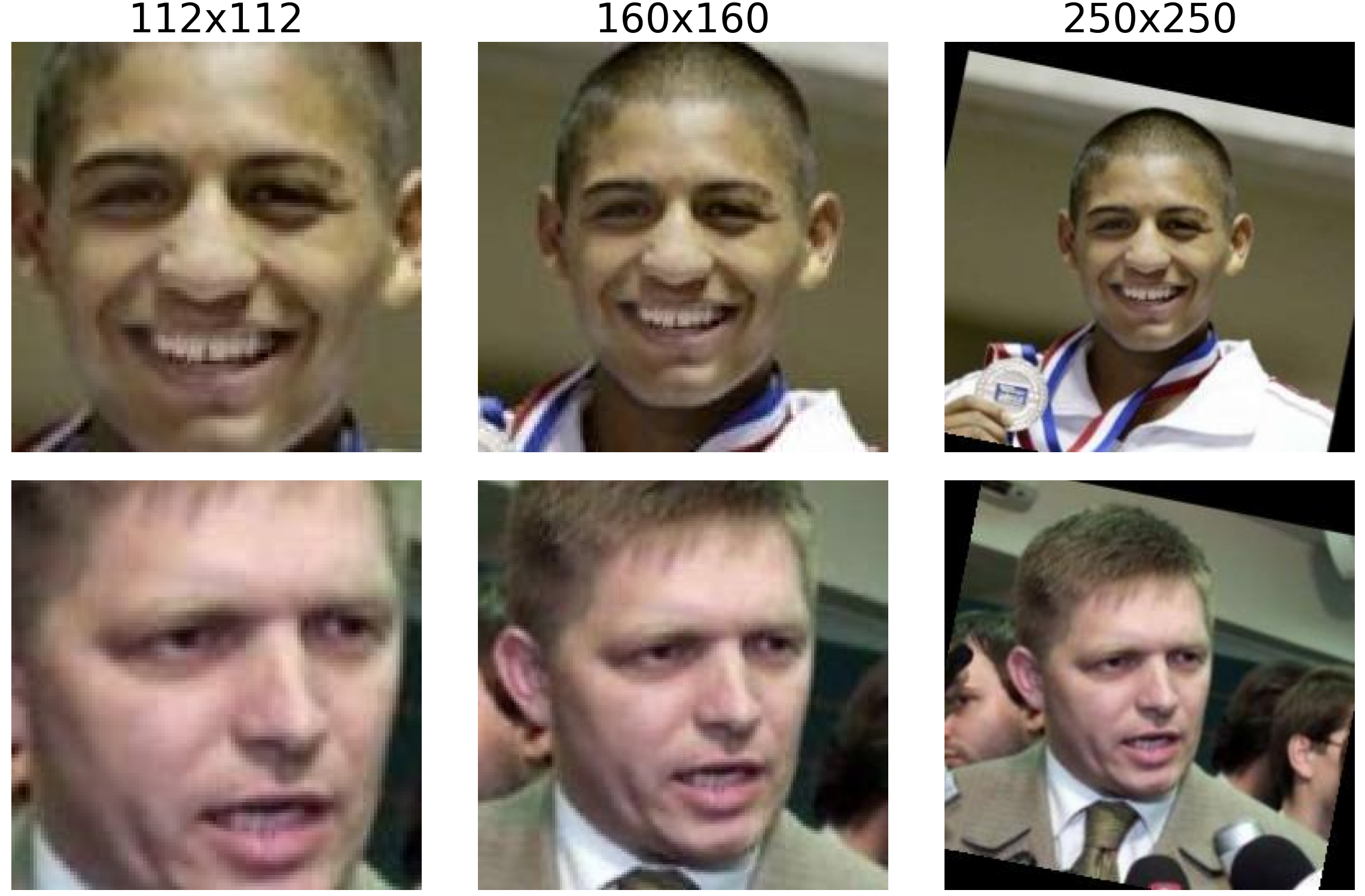}
    \captionsetup{skip=5pt} 
    \caption{Example images from the LFW dataset, showing face crops with different sizes of 112$\times$112, 160$\times$160 and 250$\times$250, as used in our experiments.}
    \label{fig:lfw-samples}
\end{figure}

\begin{table*}[ht]
\begin{minipage}{\textwidth}
\centering
\vspace{-1em}
\caption{Summary of face recognition datasets used in our experiments}
\label{tab:dataset_summary_colorfit}
\rowcolors{2}{lightgray}{white}

\begin{tabular}{>{\columncolor{lightblue}}l c c r  r}
\Xhline{1px}
\rowcolor{lightblue}
\textbf{Dataset} & \textbf{\#Subjects} & \textbf{\#Images / \#Frames} & \textbf{\#Genuine Pairs} & \textbf{\#Impostor Pairs} \\
\Xhline{1px}
LFW~\cite{cplfwtech,huang2008labeled} & 5,749 & 13,233 images & 242,257 & 87,307,271 \\
WebFace42M-Subset~\cite{zhu2021webface260m}\footnote{A subset of the larger dataset was used.} & 50,000 & 250,000 images & 100,000 & 1,249,875,000 \\
IJB-B~\cite{whitelam2017ijbb} & 1,845 & 21,798 images / 55,026 frames & 10,270 & 8,000,000 \\
IJB-C~\cite{maze2018ijbc} & 3,106 & 31,334 images / 117,542 frames & 19,557 & 15,638,932 \\
IJB-S (Still-to-Still)~\cite{kalka2018ijbs} & 202 & 1,452  images & 4,489 & 2,103,815 \\
\Xhline{1px}
\end{tabular}
\end{minipage}
\end{table*}

\section{Dataset Description}
\vspace{-0.5em}
\label{sec:dataset_description}
We used several face benchmark datasets in this work: LFW~\cite{cplfwtech}, WebFace42M-Subset~\cite{zhu2021webface260m}, IJB-B~\cite{whitelam2017ijbb}, IJB-C~\cite{maze2018ijbc}, and IJB-S~\cite{kalka2018ijbs}. See Table~\ref{tab:dataset_summary_colorfit} for details. For the LFW dataset, unlike the standard protocol, we used an exhaustive protocol (considering all possible genuine and impostor pairs) and evaluated the recognition performance on face images with different crop sizes (viz., 112$\times$112, 160$\times$160, and 250$\times$250) as shown in Figure~\ref{fig:lfw-samples}.  The WebFace42M-Subset is constructed from WebFace260M~\cite{zhu2021webface260m} by creating five disjoint sets of 10K identities, each with 5 images (50K total). All images are aligned and cropped to 112$\times$112, with no identity overlap with AdaFace or ArcFace training sets, and evaluation is done via exhaustive pairwise verification. For the IJB-B, IJB-C and IJB-S datasets, the published evaluation protocol is followed.

\section{Experimental Results}
We used three types of models: a)~thirty-five open-source foundation  models from thirteen families~\cite{jia2021scaling_align, peng2023kosmos, llava15, liu2024llavanext, chen2024internvl, wu2024deepseek_vl2, kirillov2023segment_anything, caron2021dino, oquab2023dinov2, dosovitskiy2020vit, li2022blip, li2023blip2, clip_radford2021learning, Cherti_2023_CVPR_OPENCLIP}~\footnote{We used the publicly shared weights on https://huggingface.co/} following the approach in~\cite{sony2025benchmarking}, b)~three closed-source foundation models (GPT-4o~\cite{openai2024gpt4o}, Gemini~\cite{geminiteam2025geminifamilyhighlycapable}, Grok-4~\cite{xai2025grok}), and c)~two FR models, (ArcFace~\cite{deng2019arcface} and AdaFace~\cite{kim2022adaface}), on five benchmark datasets. Model details, including input specifications, embedding dimensions, and source locations  are provided in the supplementary materials. Throughout the remainder of the paper, \textit{AdaFace (WF4M, WF12M, MS1MV2, MS1MV3)} refers to the same model architecture trained on different datasets -  WebFace4M~\cite{zhu2021webface260m}, WebFace12M~\cite{zhu2021webface260m}, MS1MV2~\cite{ms1mv2}, and MS1MV3~\cite{ms1mv3}, respectively, and by \textit{AdaFace} we refer to AdaFace~(WF12M) by default. 
The results of 1:1 verification are reported in Section~\ref{sec:verification}, fusion with open-source foundation models in Section~\ref{sec:fusion} and fusion with closed-source foundation models with explainability in Section~\ref{sec:fusion_chat_vlm}. For brevity, we report only those foundation models demonstrating performance close to the FR models. A comprehensive list of results is provided in the supplementary material for all the foundation models used in our experiments.

\subsection{Face Verification}
\label{sec:verification}
\begin{table}[t]
\centering
\caption{True Match Rate (TMR \%) at a False Match Rate (FMR) of $0.01\%$ across the three LFW variants with different crop sizes (see Figure \ref{fig:lfw-samples}).}
\rowcolors{2}{lightgray}{white}
\resizebox{\linewidth}{!}{%
\begin{tabular}{>{\columncolor{lightblue}}l c c c }
\Xhline{1px}
\rowcolor{lightblue}
& \textbf{112$\times$112} &\textbf{160$\times$160} & \textbf{250$\times$250}\\ 
\Xhline{1px}

LLaVA-1.5-7B~\cite{llava15} & 63.48 & 68.97 & 70.68 \\  
BLIP2-O-6.7B~\cite{li2023blip2} & 78.85 & 82.04 & 41.92  \\

CLIP-L-14-336~\cite{clip_radford2021learning}  & 63.49 & 68.97 & 70.68   \\

OpenCLIP-H-14~\cite{Cherti_2023_CVPR_OPENCLIP} & 64.97 & 76.83 & 81.73\\

\Xhline{1px}
ArcFace~\cite{deng2019arcface} & $98.86$ & $99.14 $& $32.88$ \\
AdaFace (WF4M)~\cite{kim2022adaface} & 98.99 & 99.24 & 56.22 \\
AdaFace (WF12M)~\cite{kim2022adaface} &99.09 & 99.27  & 77.31  \\ \Xhline{1px}
 \end{tabular}}
\label{tab:lfw_results}
\end{table}
In this Section, we compare the FR models with open-source foundation models at the 1:1 face verification task on several benchmark datasets. Further, the LFW~\cite{cplfwtech} benchmark is used for analyzing the effect of contextual cues around the face, the WebFace42M-Subset~\cite{zhu2021webface260m} for a large benchmark and IJB-B, IJB-C, IJB-S for the fusion of FR models with foundation models.  

\noindent\textbf{LFW:} To investigate the optimal operating conditions of FR and foundation models for face verification, we use the LFW~\cite{cplfwtech} dataset, which offers full-face portraits with natural backgrounds and occasional co-occurring faces—ideal for studying the influence of contextual cues. We systematically vary the cropping size as shown in Figure~\ref{fig:lfw-samples} to assess how peripheral facial and background information affects face verification performance. 

\label{sec:lfw_results_analysis}
Our experiments show that the FR models, viz., ArcFace~\cite{deng2019arcface} and AdaFace~\cite{kim2022adaface}, have a very high accuracy on tightly cropped faces. Both AdaFace and ArcFace models produced True Match Rates (TMRs) above $99.09\%$ at an FMR of $0.01\%$ in both tightly cropped 112$\times$112 and moderately over-cropped 160$\times$160 images. See Table~\ref{tab:lfw_results} for comparison. However, their performance degrades noticeably when evaluated on 250$\times$250 images containing background.

For the foundation models, the TMR of OpenCLIP-H-14~\cite{ilharco2021openclip} increased from $64.97\%$ to $81.73\%$ at an FMR of $0.01\%$ when the cropped face region was increased from 112$\times$112 to 250$\times$250,  providing contextual information around the face such as ear, hair and shoulders. Similar improvements were observed for LLaVA-1.5-7B~\cite{llava15} and CLIP-L-14-336~\cite{clip_radford2021learning}. However, for BLIP2-O-6.7B, while the performance improved when the cropped region was increased from  112$\times$112 to 160$\times$160, there was a subsequent drop in performance for the 250$\times$250 cropping. These results point to the reliance of foundation models on contextual cues and suggest that their performance on face recognition is optimal when the input face includes sufficient surrounding context, but may degrade for certain models when excessive background distracts them from identity-specific features. In contrast, FR models are typically trained on tightly cropped faces and therefore rely heavily on such inputs; their performance often drops sharply when background clutter is introduced.

\begin{table}[t]
\centering
\caption{True Match Rate (TMR \%) at False Match Rates (FMRs) of $0.01\%$ and $0.001\%$ computed over the 5  splits of disjoint identities in the WebFace42M-Subset dataset.}
\rowcolors{2}{lightgray}{white}
\begin{tabular}{ >{\columncolor{lightblue}} l c c c }
\Xhline{1px}
\rowcolor{lightblue}
 & \textbf{FMR=0.01\%} & \textbf{FMR=0.001\%}\\
\Xhline{1px}

LLaVA-1.5-7B~\cite{llava15} & 30.86 $\pm$ 0.17 & 25.54 $\pm $ 0.12  \\


BLIP2-O-6.7B~\cite{li2023blip2} & 50.21 $\pm$ 0.26 & 37.62 $\pm$ 0.44 \\
CLIP-L-14-336~\cite{clip_radford2021learning} & 30.86 $\pm$ 0.17 & 24.44 $\pm$ 0.09   \\
OpenCLIP-H-14~\cite{Cherti_2023_CVPR_OPENCLIP} & 32.58 $\pm$ 0.22 & 26.18 $\pm$ 0.16  \\ \Xhline{1px}
ArcFace (WF4M)~\cite{deng2019arcface} & 98.67 $\pm$ 0.05 & 96.97 $\pm$ 0.07 \\ 
AdaFace (WF4M)~\cite{kim2022adaface} & 98.94 $\pm$ 0.06 & 97.51 $\pm$ 0.08 \\
AdaFace (WF12M)~\cite{kim2022adaface} & 99.29 $\pm$ 0.03 & 98.31 $\pm$  0.07 \\\Xhline{1px}

\end{tabular}
\label{tab:webface42M_verification}

\end{table}
\noindent\textbf{WebFace42M-Subset:} To analyze the performance of foundation models on benchmarks with a large number of identities, we utilize the WebFace42M-Subset~\cite{zhu2021webface260m} benchmark. Here, the  FR models significantly outperform the foundation models across all FMR values in the verification task, consistently achieving a TMR over $98\%$ at an FMR of $0.01\%$ and demonstrating their superiority on large-scale benchmarks. See  Table~\ref{tab:webface42M_verification}. Among the foundation models, the BLIP2-Opt-6.7B~\cite{li2023blip2} model achieved the highest TMR of $50.21\%$ at an FMR of $0.01\%$.



\begin{table*}[ht]
\centering
\caption{Fusion of FR models with foundation models. The True Match Rate (TMR  \%) at various False Match Rates (FMRs) on the IJB-B, IJB-C, and IJB-S datasets are shown. Bold values indicate the best performance at each FMR for each dataset.}
\label{tab:ijbb_ijbc_joint_paper}
\resizebox{\textwidth}{!}{
\begin{tabular}{l|ccc|ccc|ccc}
\Xhline{1pt}
\rowcolor{lightblue}
& \multicolumn{3}{c|}{{IJB-B (TMR $\%$)}} & \multicolumn{3}{c|}{{IJB-C (TMR $\%$)}} & \multicolumn{3}{c}{{IJB-S (Still-Still) (TMR $\%$)}} \\

\hline

FMR $(\%)\Rightarrow$ & $0.0001$ & $0.001$ & $0.01$ & $0.0001$ & $0.001$ & $0.01$ & $0.0001$ & $0.001$ & $0.01$ \\
\Xhline{1px}

\multicolumn{10}{c}{\rule{0pt}{3ex}{FR Models}} \\
\Xhline{1px}

\rowcolor{domaincolor}
ArcFace (WF4M)~\cite{deng2019arcface}   & 81.27 & 89.01 & 93.91 & 83.05 & 90.58 & 95.06 & 90.29 & 91.74 & 92.45 \\

AdaFace (WF12M)~\cite{kim2022adaface}   & 72.64 & 87.5 & 94.41 & 73.17 & 88.61 & 95.38 & 87.37 & 92.23 & 92.85 \\

\Xhline{1px}
\multicolumn{10}{c}{\rule{0pt}{3ex}{Foundation Models}} \\
\Xhline{1px}

\rowcolor{foundationcolor}
CLIP-L-14-336~\cite{clip_radford2021learning}   & 49.49 & 69.46 & 86.49 & 54.48 & 72.17 & 87.33 & 5.93 & 13.52 & 29.76 \\

BLIP2-O-6.7B~\cite{li2022blip}   & 55.46 & 75.62 & 90.84 & 61.89 & 79.33 & 91.93 & 31.70 & 49.08 & 67.94 \\

\Xhline{1px}
\multicolumn{10}{c}{\rule{0pt}{3ex}{Fusion of CLIP-L-14-336 with FR Models}} \\
\Xhline{1px}

\rowcolor{fusioncolor}
ArcFace (WF4M) $+$ CLIP-L-14-336  & 80.66 & 89.44 & 94.93 & 83.86 & 91.25 & 95.86 & 77.46 & 86.95 & 91.71 \\


AdaFace (WF12M) $+$ CLIP-L-14-336 & 80.06 & 89.57 & 95.36 & 82.81 & 91.33 & 96.13 & 66.07 & 89.00 & 92.38 \\

\Xhline{1px}
\multicolumn{10}{c}{\rule{0pt}{3ex}{Fusion of BLIP2-O-6.7B with FR Models}} \\
\Xhline{1px}

\rowcolor{fusioncolor}
ArcFace (WF4M) $+$ BLIP2-O-6.7B   & 82.64 & 90.28 & 95.69 & \textbf{85.89} & 92.38 & 96.54 & \textbf{90.53} & 91.89 & 92.74 \\

AdaFace (WF12M) $+$ BLIP2-O-6.7B  & \textbf{83.31} & \textbf{90.88} & \textbf{96.05} & 85.81 & \textbf{92.78} & \textbf{96.79} & 89.04 & \textbf{92.31} & \textbf{92.92} \\

\Xhline{1px}
\end{tabular}
}
\vspace{0.5em}

\end{table*}
\noindent\textbf{IJB-B, IJB-C \& IJB-S Datasets:} Table~\ref{tab:ijbb_ijbc_joint_paper} presents the 1:1 face verification results on the IJB-B~\cite{whitelam2017ijbb} and IJB-C~\cite{kalka2018ijbs},  IJB-S (Still to Still)~\cite{maze2018ijbc} benchmarks using the vision encoders from various open-source foundation models. We observe that FR models clearly outperform the foundation models. Notably, as the FMR becomes more stringent (e.g., $0.01\%$ or lower), the performance of foundation models degrades more rapidly, falling behind that of FR models. 


\subsection{Fusion of FR Models with Foundation Models}
\label{sec:fusion}
\begin{figure*}[htbp]
\centering
\includegraphics[width=1.02\textwidth]{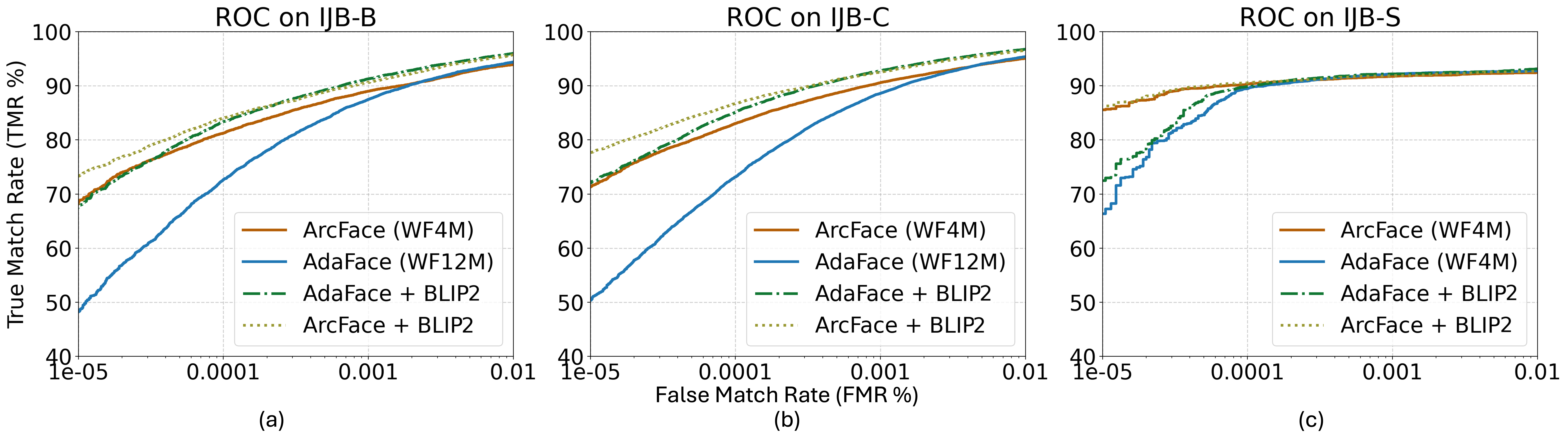}
\vspace{-1.5em}
\caption{ROC curves for the fusion of foundation model BLIP2-O-6.7B with AdaFace and ArcFace on the (a)~IJB-B, (b)~IJB-C, and (c)~IJB-S~(Still-to-Still) datasets. While the improvement is pronounced in the case of IJB-B and IJB-C, it is very modest in the case of IJB-S.}
\label{fig:ijb_roc_comparison}
\end{figure*}
As previously observed in Section~\ref{sec:lfw_results_analysis}, FR models require tightly cropped facial images, while foundation models benefit from the inclusion of contextual information in loosely cropped face images. On challenging benchmarks such as IJB-B, IJB-C, and IJB-S where facial regions are tightly localized (the dataset is already cropped), even AdaFace and ArcFace models exhibit performance limitations. To integrate local facial details from FR models with contextual cues from foundation models, we perform score-level fusion as shown in Figure~\ref{fig:fusion_diagram}. For example, when we combine the best-performing foundation model (BLIP2-O-6.7B) with AdaFace, the TMR improves from 72.64\% to 83.31\% on IJB-B and from 73.17\% to 85.81\% on IJB-C at an FMR of $0.0001\%$. As can be seen in Table~\ref{tab:ijbb_ijbc_joint_paper}, these gains are more pronounced at lower FMRs. For the IJB-S Still-to-Still benchmark, the fusion yields only modest overall gains. The Receiver Operating Characteristic (ROC) curves in Figure~\ref{fig:ijb_roc_comparison} highlight notable improvements at lower FMRs, where combining foundation models with AdaFace enhances the TMRs. Figure~\ref{fig:visual_abstract} illustrates cases from the IJB-B dataset where the AdaFace model fails, but fusion with the foundation model (BLIP2-O-6.7B) results in the correct decision. This suggests that foundation models are extracting complementary cues that improve face recognition in challenging cases.

\subsection{Explainability Using Foundation Models}
\label{sec:fusion_chat_vlm}
For explainability of face comparison outcomes, we used three closed-source foundation models: GPT-4o~\cite{openai2024gpt4o,openai2023gpt4}, Gemini~\cite{geminiteam2025geminifamilyhighlycapable} and Grok-4~\cite{xai2025grok}.  We examined whether these closed-source foundation models (accessed via API)\footnote{As per the agreement clauses of OpenAI, Google, and xAI (August 2025), the data will not be used to train or improve models (unless explicitly opted‑in).} could generate human-understandable explanations for face verification outcomes. The goal of this analysis was not to assess their recognition accuracy, but to evaluate their ability to articulate visual reasoning consistent with facial similarity judgments made by a FR model, namely AdaFace~\cite{kim2022adaface}. While DeAndres-Tame et al.~\cite{deandres2024good} conducted explainability experiments using GPT-4~\cite{openai2023gpt4} and compared its decisions with those of AdaFace, our work systematically evaluates multiple prompt formulations and analyzes explainability provided by GPT-4o~\cite{openai2024gpt4o} and Grok-4 performance on borderline and challenging face pairs across two datasets.  We experimented with several versions of models from the OpenAI library, as well as Gemini and Grok-4, and found that GPT-4o consistently provided the best performance in terms of explainability. While Grok-4 also showed good performance compared to GPT-4o, Gemini failed to deliver meaningful explanations.

\textbf{LFW:} Experimenting with a number of different prompts at the beginning, we observed that GPT-4o consistently highlighted meaningful facial cues—such as face geometry of the “face triangle” (eyes, nose, mouth), eyebrow shape, skin texture cues (e.g., wrinkles, smile lines), hairline and forehead contours, and subtle distinctions like eyelid shape or cheekbone structure, eyebrow thickness and placement, nose bridge and tip shape, and jawline structure—even correcting some of AdaFace's errors in some cases. Based on these observations, we selected \textbf{three} representative prompts to evaluate its reasoning across pairs.  Note that the wording in \textit{Prompt-1} was modified in the case of a non-match by AdaFace. To reduce bias from prompt phrasing or filenames, we used a neutral prompt and randomized filenames.

\begin{tcolorbox}[promptbox, title= Examples of the prompts used in our experiments, label={box:prompts}]
    \textbf{Prompt-1:} \textbf{AdaFace} gives a similarity score of \textbf{0.3021} for this face image pair. This is, therefore, considered to be a match. Could you examine these two face images and explain why they are  of the same person?
    
    \vspace{4pt} \textbf{Prompt-2:} \textbf{A face recognition model} gives a similarity score of \textbf{0.3021} for this face image pair. Could you examine these two face images and explain why?
    
    \vspace{4pt} \textbf{Prompt-3:} Please examine these two face images. Are they of the same person or not? Explain your reasoning.
\end{tcolorbox}

\begin{table}[!htbp]
\captionsetup{skip=2pt} 
\caption{Accuracy of GPT-4o explanations under three prompt phrasings (mentioned  in Section~\ref{sec:fusion_chat_vlm}). For each prompt we evaluated \textbf{200} LFW image pairs—\textbf{50 in each category} (FNM, FM, TNM, TM) chosen from AdaFace’s correctly and incorrectly classified pairs.  Numbers denote “\#correct responses~/~\#total pairs” for that category.}
\centering
\small
\setlength{\tabcolsep}{6pt}
\renewcommand{\arraystretch}{1.25}
\begin{tabular}{l c c c}
\rowcolor{headerblue!80!black}
\textcolor{white}{\textbf{Pair Type}} &
\textcolor{white}{\textbf{Prompt 1}} &
\textcolor{white}{\textbf{Prompt 2}} &
\textcolor{white}{\textbf{Prompt 3}} \\ \hline
False Non-Match (FNM) & \cellcolor{warnred}{$10 / 50$} & \cellcolor{warnred}$38 / 50$ & \cellcolor{goodgreen}$50 / 50$ \\
False Match (FM)     & \cellcolor{warnred}{$6 / 50$}     & \cellcolor{warnred}$12 / 50$   & \cellcolor{goodgreen}$50 / 50$ \\
True Non-Match (TNM)  & \cellcolor{goodgreen}$50 / 50$                       & \cellcolor{goodgreen}$50 / 50$ & \cellcolor{goodgreen}$50 / 50$ \\
True Match (TM)      & \cellcolor{goodgreen}$50 / 50$                         & \cellcolor{goodgreen}$50 / 50$   & \cellcolor{goodgreen}$50 / 50$ \\ \hline
\rowcolor{headerblue!20}
\textbf{Total correct} & $116 / 200$                     & $150 / 200$ & $200 / 200$ \\
\end{tabular}
\label{tab:prompt_accuracy}
\end{table}

\begin{table*}[ht]
\centering
\caption{GPT-4o explanations for LFW and IJB-S misclassified pairs show prompt phrasing strongly impacts reasoning, with \textit{Prompt-3} yielding the most accurate reasoning.}
\vspace{-0.2em}
\label{tab:lfw-ijbs-images-responses}
\resizebox{\linewidth}{!}{%
\begin{tabular}{|p{10.3cm}|p{10.3cm}|}
\multicolumn{2}{c}{\textbf{Responses from GPT-4o with Prompts in Section~\ref{sec:fusion_chat_vlm} on LFW~\cite{cplfwtech} Samples}}  \\ \Xhline{1px}
\rowcolor{lightblue}
\multicolumn{1}{|c}{\textbf{False Match (FM)}} & \multicolumn{1}{|c|}{\textbf{False Non-Match (FNM)}}   \\\Xhline{1px}
\makebox[\linewidth][c]{
  \begin{minipage}{0.5\textwidth}
    \centering
    \includegraphics[width=0.25\textwidth]{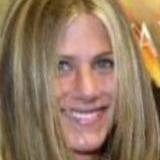}
    \hspace{1mm}
    \includegraphics[width=0.25\textwidth]{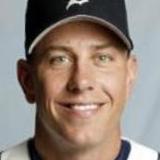}
  \end{minipage}
}
&
\makebox[\linewidth][c]{
 \begin{minipage}{0.5\textwidth}
    \centering
    \includegraphics[width=0.25\textwidth]{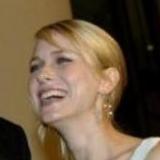}
    \hspace{1mm}
    \includegraphics[width=0.25\textwidth]{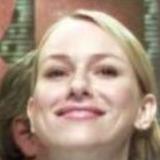}
\end{minipage}}\\
\hline
\textbf{Prompt 1 Response:} \textbf{AdaFace explains the match} by aligning \textbf{eye spacing, nose shape, and lip proportions}, ignoring superficial differences (e.g., \textbf{hair, lighting, cap}). These consistent \textbf{craniofacial features} justify the \textcolor{red}{\textbf{high similarity score}}. &     \textbf{Prompt 1 Response:} AdaFace detects misalignments in \textbf{eyes, nose, jawline, and face proportions}. These structural differences result in a \textbf{low similarity score}, leading to a confident \textcolor{red}{\textbf{non-match}}. \\\hline

\textbf{Prompt 2 Response:} The model relies on stable biometric traits like \textbf{eye shape, face contour, and nose geometry}. Despite differences in \textbf{style and gender}, the \textbf{overlap in deep features} leads to a \textcolor{red}{\textbf{match}}. Mentions risk of false positives. & \textbf{Prompt 2 Response:} \textbf{Pose and expression mismatches} distort key facial landmarks like \textbf{eye spacing and jaw contour}, making alignment hard. The model reports a \textcolor{red}{\textbf{non-match}} due to lack of overlap.  \\\hline

\textbf{Prompt 3 Response:} Despite similar smiles, clear differences in \textbf{jawline, brow shape, nose width, and eye spacing} point to \textbf{two different identities}. Declares a confident \textcolor{blue}{\textbf{non-match}} based on \textbf{structural mismatches}. & \textbf{Prompt 3 Response:} Despite \textbf{minor pose differences}, strong consistency in \textbf{eye shape, smile, facial structure}, and even details like \textbf{earring position} suggest these are the \textbf{same person}. Declares a confident \textcolor{blue}{\textbf{match}}.   \\ \Xhline{1.5px}
\multicolumn{2}{c}{}   \\
\multicolumn{2}{c}{\textbf{Responses from GPT-4o with Prompts in Section~\ref{sec:fusion_chat_vlm} on IJB-S~\cite{kalka2018ijbs} Samples }}  \\ \Xhline{1px}
\rowcolor{lightblue}
\multicolumn{1}{|c|}{\textbf{False Match (FM)}} & \multicolumn{1}{|c|}{\textbf{False Non-Match (FNM)}}   \\\Xhline{1px}
\makebox[\linewidth][c]{
 \begin{minipage}{0.5\textwidth}
    \centering
    \includegraphics[width=0.25\textwidth]{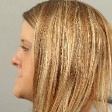}
    \hspace{1mm}
    \includegraphics[width=0.25\textwidth]{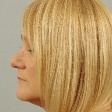}
\end{minipage} }
& 
\makebox[\linewidth][c]{
 \begin{minipage}{0.5\textwidth}
    \centering
    \includegraphics[width=0.25\textwidth]{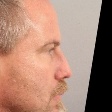}
    \hspace{1mm}
    \includegraphics[width=0.25\textwidth]{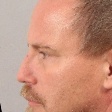}
\end{minipage} }
\\

\Xhline{1px}
\textbf{Prompt 1 Response:} Both images are in a \textbf{strict left-profile pose} with \textbf{similar hair, skin tone, and facial structure}—notably the \textbf{nose, chin, and jawline}. These overlapping \textbf{geometric cues} enable the model to align features well. Despite lacking frontal details, the \textbf{global consistency} across both images leads to a confident \textcolor{red}{\textbf{match decision}}. & \textbf{Prompt 1 Response:} Despite the low similarity score of \textbf{0.2194}, the pair shares \textbf{consistent facial structure} (forehead slope, nose bridge, chin), \textbf{matching facial hair pattern}, \textbf{identical skin tone and texture}, and \textbf{similar hairline and color}. The strict side profile limits landmark visibility, likely causing AdaFace to underestimate the match. Visually, these images are \textcolor{blue}{\textbf{likely of the same person}}.
   \\ \hline 
\textbf{Prompt 2 Response:} The similarity score of \textbf{0.7915} suggests a strong match despite the \textbf{challenging profile view}. Both faces have \textbf{closely aligned jawline, nose shape, and forehead slope}, along with \textbf{similar blonde hair} and \textbf{identical lighting conditions}. These \textbf{consistent geometric and contextual cues} result in high embedding similarity and \textcolor{red}{\textbf{support the match decision}}. & 
\textbf{Prompt 2 Response:} Although AdaFace assigns a low similarity score of \textbf{0.2194}, both images exhibit \textbf{high visual similarity}—including \textbf{matching hair color and balding pattern}, \textbf{facial hair}, \textbf{eyebrow shape}, \textbf{skin tone}, and \textbf{ear structure}. The \textbf{side profile view}, \textbf{lighting variation}, and \textbf{slight rotation} likely degraded the embedding quality. These factors suggest a \textbf{false negative}, and the pair is \textcolor{blue}{\textbf{likely of the same person}} despite the model’s decision.  \\ \hline

\textbf{Prompt 3 Response:} Both images are in \textbf{strict profile view}, limiting visible facial features. While they share \textbf{similar blonde hair}, differences in \textbf{hair parting and hairline} are evident. The \textbf{nose shape, jawline}, and \textbf{facial roundness} also differ noticeably. Given the \textbf{limited identity cues} and these structural mismatches, these images are likely \textcolor{blue}{\textbf{not of the same person}}. & 
\textbf{Prompt 3 Response:} Despite minor \textbf{lighting and head tilt differences}, the two \textbf{left-profile images} show \textbf{high consistency} in \textbf{forehead slope, nose shape, chin contour, receding hairline}, and \textbf{facial hair distribution}. \textbf{Skin texture and tone} are also matched. These overlapping \textbf{biometric features} indicate that the two images are \textcolor{blue}{\textbf{of the same individual}}.   \\ \hline

\Xhline{1px}
    \end{tabular}}
\end{table*}
A key insight was that prompt wording~\cite{sahoo2024systematic} significantly affected both performance and explanation quality. When \textit{Prompt-1} explicitly mentioned ``AdaFace”, GPT-4o often apparently anchored its explanation around AdaFace’s known criteria—justifying decisions using concepts such as embedding similarity and ``face-print” geometry. To control for this, we repeated the analysis with \textit{Prompt-2}, replacing ``AdaFace” with a generic phrase like \textit{a face recognition model}. This neutral phrasing led to slight improvements in explanation quality, though the model still referenced typical decision factors used by FR systems mentioned earlier with \textit{Prompt-1}.

The best explanations were obtained with \textit{Prompt-3}, which did not have any reference to the model and score. \textbf{``Please examine these two face images. Are they of the same person or not? Explain your reasoning''}. Using this fully neutral prompt, GPT-4o correctly classified all 200 image pairs across the four ground-truth categories (TM, TNM, FM, FNM), as shown in Table~\ref{tab:prompt_accuracy}. Notably, the explanations remained accurate even when AdaFace failed—particularly in False Match and False Non-Match cases. For example, although AdaFace misclassified some pairs due to background clutter, occlusion, or poor detection crops, GPT-4o still provided correct visual interpretations based on observable identity cues although it appears gender cues were not used in those decisions. It pointed out when faces were cropped differently or when one image contained a different person entirely due to failed detection. Even in low-confidence AdaFace matches (e.g., scores around 0.21–0.24), GPT-4o accurately highlighted subtle geometric alignments in facial structure and explained why these were (or were not) sufficient for a genuine match. Results in Table~\ref{tab:prompt_accuracy} show:

\begin{itemize}
\item For True Match (TM) and True Non-Match (TNM) pairs, GPT-4o consistently produced correct decisions with detailed and accurate explanations.
\item For False Match (FM) pairs, GPT-4o frequently reported mismatch in facial features (e.g., differing eye spacing, jaw width, or skin texture), yet aligned its final decision with that of AdaFace’s similarity score—incorrectly labeling the pair as a match despite having recognized the discrepancies.
\item For False Non-Match (FNM) pairs, while some explanations and final classifications were incorrect, we still observed useful information in the textual explanation. In several cases, GPT-4o correctly described critical identity-related features—such as consistent face shape, eye spacing, cheek and jaw contour, nose bridge structure, and even fine texture patterns like wrinkles or stubble—even though it ultimately followed AdaFace's incorrect non-match decision (a subset of these pairs is presented in Table \ref{tab:lfw-ijbs-images-responses}). This highlights the potential of foundation models to produce textual explanation even in challenging cases.
\end{itemize}

These results underscore that foundation models like GPT-4o can generate detailed and interpretable reasoning grounded in facial features independent of the embedding space or match score. Even when AdaFace failed (e.g., in False Match/False Non-Match pairs), the explanations of GPT-4o  aligned well with human facial analysis, reinforcing its value for post-hoc explainability in face recognition systems.

\begin{figure}[h]
\centering
\setlength{\tabcolsep}{1pt}  
\begin{tabular}{cccc}
\includegraphics[width=0.22\linewidth]{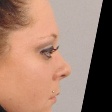} &
\includegraphics[width=0.22\linewidth]{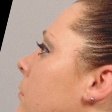} &
\includegraphics[width=0.22\linewidth]{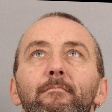} &
\includegraphics[width=0.22\linewidth]{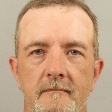} \\
\multicolumn{2}{c}{\scriptsize \textbf{Genuine — AdaFace \xmark, GPT-4o \checkmark}} &
\multicolumn{2}{c}{\scriptsize \textbf{Impostor — AdaFace \xmark, GPT-4o \checkmark}} \\
\end{tabular}
\vspace{-6pt}
\captionsetup{skip=5pt} 
\caption{GPT-4o correctly classified these ``edge" cases from the IJB-S dataset that were close to the threshold corresponding to an FMR of $0.01\%$, where AdaFace failed.}
\label{fig:adaface-failure}
\end{figure}

\noindent\textbf{IJB-S:} To assess face recognition performance under more challenging and realistic conditions, we extended our experiments to the IJB‑S~\cite{kalka2018ijbs} dataset and utilized GPT-4o alongside another recent model, Grok-4. We focused on the most difficult genuine and impostor pairs identified by AdaFace. We selected 2,000 genuine pairs with low AdaFace match scores and 2,000 impostor pairs with high match scores on  the IJB-S dataset. As shown in Table~\ref{tab:ijbs_borderline_expB}, both GPT‑4o and Grok‑4 demonstrate improved decision-making on these hard samples compared to AdaFace, with an overall accuracy of $78.80\%$ for GPT‑4o and $58.60\%$ for Grok‑4, versus $41.98\%$ for AdaFace. In particular, GPT‑4o achieved high accuracy on both genuine and impostor cases, despite refusing to score $221$ pairs due to uncertainty (we labeled this as GPT-4o errors). Grok‑4 produced no refusals and achieved competitive performance with detailed explanations (an example is shown in Table \ref{tab:grok_false_positive_table}). We also conducted a broader evaluation by selecting all available 4,489 genuine pairs and randomly sampling 20,000 impostor pairs from the IJB-S dataset. We applied score-level fusion between GPT-4o/Grok-4 and AdaFace on this larger set; however, the fusion did not lead to any improvement in verification performance. To enable score-level fusion, we added a condition to \textit{Prompt-3} asking the model to \textit{``give a similarity score between 0--100''} along with its decision. These results suggest that foundation models, even without explicit biometric training, are capable of making strong decisions on difficult samples and providing useful explanations to support their predictions. Example pairs are shown in Figure~\ref{fig:adaface-failure}.

\definecolor{lightgray}{gray}{0.95}
\definecolor{lightblue}{RGB}{230,240,255}
\definecolor{lightgreen}{RGB}{235,255,235}

\begin{table}[htbp]
\centering
\small
\captionsetup{skip=2pt}
\caption{Challenging genuine and impostor pairs as assessed by AdaFace  on the IJB‑S  dataset. GPT-4o and Grok‑4 have better recognition accuracy than AdaFace on these pairs.}
\rowcolors{2}{white}{lightgray}

\resizebox{\linewidth}{!}{%
\begin{tabular}{l | c c c}
\Xhline{1px}
\rowcolor{lightblue}
 \textbf{\#Pairs}&\textbf{ AdaFace~\cite{kim2022adaface} }& \textbf{GPT‑4o~\cite{openai2024gpt4o}} & \textbf{Grok‑4}~\cite{xai2025grok} \\
\Xhline{1px}
Genuine / Impostor & 2,000/2,000 & 2,000/2,000 & 2,000/2,000 \\
\rowcolor{lightgreen}
Correctly Classified Genuine & 1,679 (83.95\%) & 1,613 (80.65\%) & 1,826 (91.30\%) \\
\rowcolor{lightgreen}
Correctly Classified Impostor & 0 & 1,365 (68.25\%) & 518 (25.90\%) \\
\Xhline{1px}
\rowcolor{green}
Overall Accuracy & 41.98\% & 74.45\% & 58.60\% \\
\Xhline{1px}
\end{tabular}}
\label{tab:ijbs_borderline_expB}
\end{table}

\begin{table*}[htbp]
\caption{False Match (FM) cases due to AdaFace on the LFW dataset analyzed by Grok‑4. Images and full explanation shown below. The response for these images by GPT‑4o is shown in Table~\ref{tab:lfw-ijbs-images-responses} and supplementary materials.}
\label{tab:grok_false_positive_table}
\resizebox{\linewidth}{!}{%
\begin{tabular}{|p{0.35\linewidth}|p{0.68\linewidth}|}
\Xhline{1px}
\rowcolor{lightblue}
\multicolumn{1}{|c|}{\textbf{False Match (FM)}} & \multicolumn{1}{c|}{\textbf{False Match (FM)}}  \\\Xhline{1px}
\makebox[\linewidth][c]{
 \begin{minipage}{0.5\textwidth}
    \centering
    \includegraphics[width=0.25\textwidth]{images/LFW_FP/steve_0001.jpg}
    \hspace{1mm}
    \includegraphics[width=0.25\textwidth]{images/LFW_FP/jen_0004.jpg}
\end{minipage} } 
&
\makebox[\linewidth][c]{
 \begin{minipage}{0.5\textwidth}
    \centering
    \includegraphics[width=0.25\textwidth]{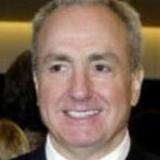}
    \hspace{1mm}
    \includegraphics[width=0.25\textwidth]{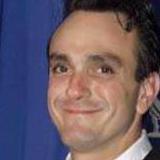}
\end{minipage} } \\\hline
\noindent\textbf{Grok‑4 Response:} A detailed visual and anthropometric analysis confirms this is a \textbf{false positive match} between two \textbf{different individuals}. While both faces exhibit \textbf{some proportional alignments}—such as \textbf{interocular distance} (30–35\% of face width), a \textbf{leptorrhine nasal index} (0.6–0.7), and a \textbf{facial index} near 1.3—these are outweighed by \textbf{gender-dimorphic features}. The left image shows a \textbf{pronounced jawline}, \textbf{thicker brow ridges}, and \textbf{short dark hair wearing a cap}, consistent with \textbf{male androgenic traits}, while the right has \textbf{softer facial contours}, \textbf{lighter long hair}, and \textbf{subtler brow arches}, matching \textbf{female estrogenic norms}. Additional differences in \textbf{skin texture}, \textbf{hair follicle density}, and \textbf{attire context} further support identity mismatch. No digital tampering is evident. Although the two faces share a few aligned ratios (e.g., golden ratio approximations), the \textbf{overall facial gestalt} and \textbf{dimorphic markers} diverge significantly. This is corroborated by the low similarity score (\textbf{0.2374}), which reflects substantial \textbf{embedding vector divergence}, leading to the conclusion that they are \textbf{not the same person} under both biometric and biological scrutiny. & \noindent\textbf{Grok‑4 Response:} Based on a detailed visual and anthropometric analysis, it is concluded that the two provided images depict \textbf{distinct individuals}, not the same person. The \textbf{cumulative differences} in facial morphology, pigmentation, and proportional metrics \textbf{significantly outweigh the similarities}, indicating \textbf{unique identities} without any evidence of aging, alterations, or imaging artifacts that could account for the observed variation. While both faces share some \textbf{broad categorical traits}, such as \textbf{mesoprosopic oval facial shapes} (facial index $\sim$1.25--1.3), \textbf{similar interocular distances} (about 32\% of face width), and \textbf{moderate labial proportions with midline symmetry} (deviation $<$5\%), these similarities are superficial. Key \textbf{morphological divergences} include \textbf{hair pigmentation and texture}—the left image features \textbf{lighter, straighter blondish-grey hair} with a \textbf{receding hairline} (forehead exposure $\sim$40\%), while the right image shows \textbf{darker, curlier brown-black hair} with a \textbf{lower and denser implantation line} (exposure $\sim$20\%). Their \textbf{nasal structures} differ as well: the left has a \textbf{narrower, straighter bridge} (nasal index $\sim$0.55) with a refined tip, compared to the right's \textbf{broader, subtly convex nose} (index $\sim$0.65) and bulbous apex. The \textbf{ocular regions} also vary, with the left showing \textbf{narrower palpebral fissures} ($\sim$25\% of face width) and \textbf{hooded lids}, versus the right’s \textbf{wider eyes} ($\sim$28\%) and open arches. Further, the \textbf{perioral features} show distinct contrast: the left has \textbf{thinner lips} and \textbf{subtle nasolabial folds}, while the right exhibits \textbf{fuller lips} and a \textbf{right-sided asymmetric smirk} (elevation $\sim$10\% higher). Differences in \textbf{skin texture and apparent age} are also significant—the left appears older (mid-50s to 60s) with \textbf{finer pores} and \textbf{orbital hollowing}, whereas the right appears younger (40s to early 50s) with a \textbf{smoother, tauter dermis}. Even \textbf{crow’s feet patterns} differ by $\sim$20\%. Contextual elements such as \textbf{uniform suit collars} and \textbf{similar blue backgrounds} provide no strong linkage, and analysis confirms the \textbf{absence of image manipulation or artifacts}. Guided by \textbf{anthropometric standards} (e.g., \textbf{Farkas norms}) and \textbf{holistic facial recognition principles}, these \textbf{non-overlapping biometric features} outweigh general similarities. The conclusion, drawn with \textbf{high confidence}, is that these two faces represent \textbf{two separate individuals}. \\ \Xhline{1px}
\end{tabular}}
\end{table*}

\section{Summary}
\vspace{-0.5em}

In this work, we compared the performance of domain-specific FR models with foundation models. Based on a series of experiments, we draw the following conclusions:
\begin{enumerate}
\itemsep0em
    \item FR models are vastly superior to the foundation models on all the datasets considered in this work for FR tasks. 
    \item The performance of foundation models was observed to improve when the face region was over-segmented, i.e., \textit{loosely cropped}, thereby suggesting the importance of broader contextual clues in the case of foundation models. FR models, on the other hand, were observed to degrade in performance when face images were over-segmented. 
    \item The score-level fusion of FR models with foundation models resulted in an improvement in recognition accuracy thereby suggesting that complementary information was being encoded by these models.
    \item Foundation models provide explainability via textual reasoning when comparing two face images and, in some instances, resolve ambiguous decisions made by the FR models. Future work will explore deploying these models in operational FR systems to enhance both decision reliability and explainability.
\end{enumerate}

\FloatBarrier

\newpage
{
    \small
    \balance
    \bibliographystyle{ieeenat_fullname}
    \bibliography{main}
}

\clearpage
\newpage

\end{document}